\documentclass[letterpaper]{article} % DO NOT CHANGE THIS
\usepackage{aaai2026}  % DO NOT CHANGE THIS
\usepackage{times}  % DO NOT CHANGE THIS
\usepackage{helvet}  % DO NOT CHANGE THIS
\usepackage{courier}  % DO NOT CHANGE THIS
\usepackage[hyphens]{url}  % DO NOT CHANGE THIS
\usepackage{graphicx} % DO NOT CHANGE THIS
\def\UrlFont{\rm}  % DO NOT CHANGE THIS
\usepackage{natbib}  % DO NOT CHANGE THIS AND DO NOT ADD ANY OPTIONS TO IT
\usepackage{caption} % DO NOT CHANGE THIS AND DO NOT ADD ANY OPTIONS TO IT
\usepackage{amsmath,amssymb,amsfonts}
\usepackage{booktabs}
\usepackage{multirow}
\usepackage{makecell}

\usepackage{algorithm}
\usepackage{algorithmic}

\usepackage{newfloat}
\usepackage{listings}
\DeclareCaptionStyle{ruled}{labelfont=normalfont,labelsep=colon,strut=off} % DO NOT CHANGE THIS
\floatstyle{ruled}
\newfloat{listing}{tb}{lst}{}
\floatname{listing}{Listing}
\newcommand{\equalcontrib}{\textsuperscript{*}}

\title{GEMA-Score: Granular Explainable Multi-Agent Scoring Framework for Radiology Report Evaluation}

\author{
Zhenxuan Zhang\textsuperscript{\rm 1}\equalcontrib,
KinHei Lee\textsuperscript{\rm 1}\equalcontrib,
Peiyuan Jing\textsuperscript{\rm 1},
Weihang Deng\textsuperscript{\rm 1},
Huichi Zhou\textsuperscript{\rm 1},
Zihao Jin\textsuperscript{\rm 1},
Jiahao Huang\textsuperscript{\rm 1},
Zhifan Gao\textsuperscript{\rm 2},
Dominic C. Marshall\textsuperscript{\rm 3},
Yingying Fang\textsuperscript{\rm 1}\thanks{Corresponding authors.},
Guang Yang\textsuperscript{\rm 1}\footnotemark[2]
}

\affiliations{
\textsuperscript{\rm 1}Department of Bioengineering, Imperial College London, UK\\
\textsuperscript{\rm 2}School of Biomedical Engineering, Sun Yat-sen University, China\\
\textsuperscript{\rm 3}Department of Surgery \& Cancer, Imperial College London, UK\\
\texttt{\{z.zhenxuan24, k.lee24, peiyuan.jing22, weihang.deng24, h.zhou24, z.jin23, j.huang21, g.yang\}@imperial.ac.uk}\\
\texttt{gaozhifan@gmail.com, dominic.marshall12@imperial.ac.uk, y.fang@imperial.ac.uk}
}

\begin{document}
\maketitle
\begin{abstract}
Automatic medical report generation has the potential to support clinical diagnosis, reduce the workload of radiologists, and demonstrate potential for enhancing diagnostic consistency. However, current evaluation metrics often fail to reflect the clinical reliability of generated reports. Early overlap-based methods focus on textual matches between predicted and ground-truth entities but miss fine-grained clinical details (e.g., anatomical location, severity). Some diagnostic metrics are limited by fixed vocabularies or templates, reducing their ability to capture diverse clinical expressions. LLM-based approaches further lack interpretable reasoning steps, making it hard to assess or trust their behavior in safety-critical settings. These limitations hinder the comprehensive assessment of the reliability of generated reports and pose risks in their selection for clinical use. Therefore, we propose a Granular Explainable Multi-Agent Score (GEMA-Score) in this paper,  which conducts both objective quantification and subjective evaluation through a large language model-based multi-agent workflow. Our GEMA-Score parses structured reports and employs stable calculations through interactive exchanges of information among agents to assess disease diagnosis, location, severity, and uncertainty. Additionally, an LLM-based scoring agent evaluates completeness, readability, and clinical terminology while providing explanatory feedback. Extensive experiments validate that GEMA-Score achieves the highest correlation with human expert evaluations on a public dataset, demonstrating its effectiveness in clinical scoring (Kendall coefficient = $0.69$ for ReXVal dataset and Kendall coefficient = $0.45$ for RadEvalX dataset). The anonymous project demo is available at: \url{https://github.com/Zhenxuan-Zhang/GEMA_score}.
\end{abstract}


\section{Introduction}
\begin{figure*}[t]
\centerline{\includegraphics[width=2.1\columnwidth]{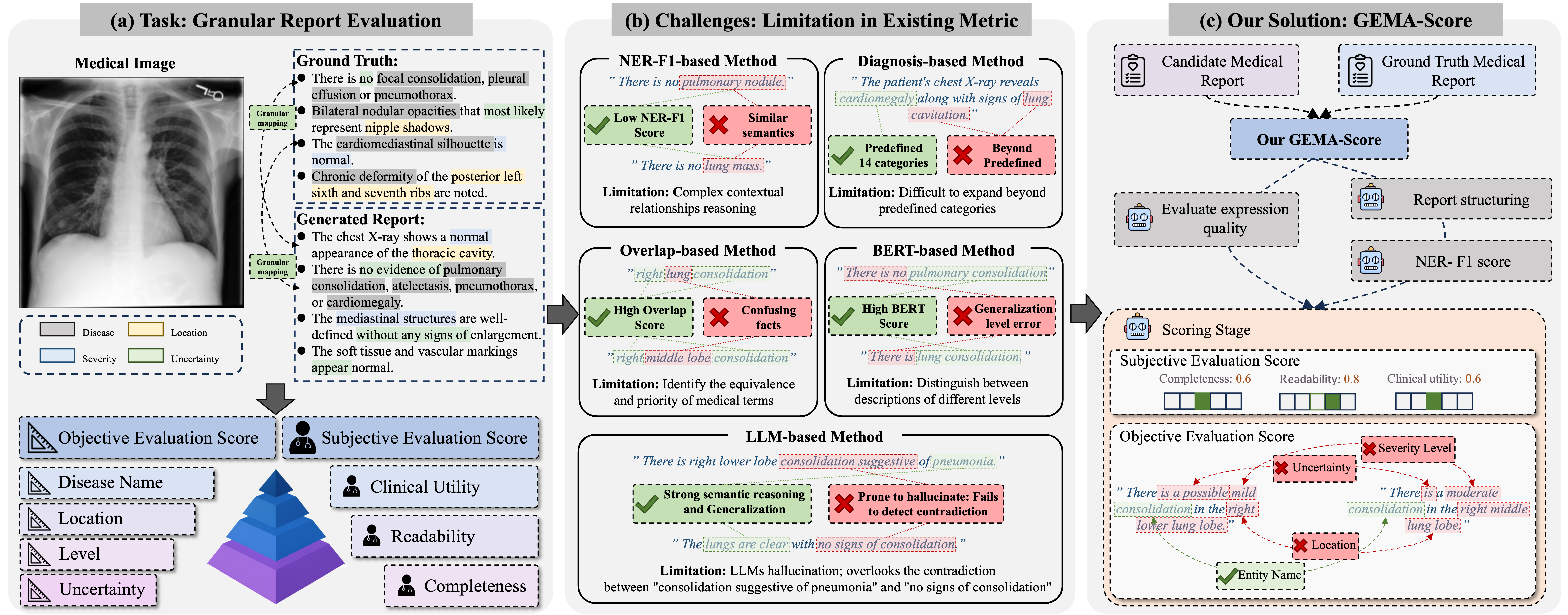}}
\caption{Motivation of our GEMA-Score. (a) The task of evaluating generated medical reports using objective and subjective metrics. (b) Limitations of existing evaluation metrics, including NER-F1, overlap-based, and BERT-based methods. (c) The proposed GEMA-Score provides a comprehensive assessment of generated reports. }
\label{fig1}
\end{figure*}

Automatic medical report generation (AMRG) has the potential to support clinical diagnosis and alleviate radiologists' workload. Many studies have explored methods to automatically generate high-quality diagnostic reports~\cite{r2gen,chexagent,radfm}. With the proliferation of AMRG models, evaluating report quality is crucial to ensure accuracy, manage patient risks and maintain clinical reliability. Traditional metrics often fail to capture clinical practicality, as report quality depends not only on covering key medical information but also on accurate and granular expression~\cite{meteor,rouge,bleu}. For instance, chest X-ray reports must indicate lesion type (e.g., atelectasis, pleural effusion), location (e.g., left lower lobe, bilateral), severity (e.g., mild, extensive), and uncertainty (e.g., probable, possible). Missing these details can lead to misdiagnosis and poor clinical decisions~\cite{Yu2022.08.30.22279318}. Additionally, clear and logical language is vital to avoid patient misunderstanding. Therefore, a new fine-grained and explainable evaluation metric is needed.

Existing evaluation metrics can be broadly grouped into five categories (Fig.~\ref{fig1}): overlap-based, BERT-based, NER-F1-based, diagnostic-based, and LLM-based methods \cite{bleu,rouge,bertscore,radgraph,smit2020chexbertcombiningautomaticlabelers,radfm,green,doclens,ratescore}. These metrics differ in linguistic granularity, clinical relevance, and interpretability. Overlap-based metrics (e.g., BLEU~\cite{bleu}, ROUGE-L~\cite{rouge}) rely on exact n-gram matches but fail to handle synonyms, minor phrasing differences, or contradictions \cite{10.1145/3442188.3445909,Yu2022.08.30.22279318}. BERT-based metrics (e.g., BERTScore~\cite{bertscore}) use contextual embeddings to assess semantic similarity, but often overestimate scores and miss clinical correctness. NER-F1-based metrics (e.g., RadGraphF1~\cite{radgraph}) focus on medical entities but ignore synonymy and entity correctness. Diagnostic-based metrics (e.g., CheXbert~\cite{smit2020chexbertcombiningautomaticlabelers} and Radbert~\cite{radbert}) classify predefined conditions, but are constrained by label scope and annotation costs \cite{radfm}. LLM-based metrics (e.g., GREEN~\cite{green}, DocLens~\cite{doclens}, RaTE~\cite{ratescore}) aim to offer holistic evaluation but often lack interpretability and condition-level attribution \cite{gu2024probabilistic,pal2023med}. 
Given the limitations of existing metrics, RadCliQ~\cite{Yu2022.08.30.22279318} integrates human evaluation criteria and correlates well with expert scores. Yet it remains coarse-grained and annotation-heavy.

Current evaluation faces several fundamental challenges that undermine its reliability and clinical utility. First, most metrics fail to account for clinically relevant variations in expression, such as synonyms (e.g., "opacity" vs. "infiltrate"), uncertainty modifiers ("likely", "possible"), or severity descriptors ("mild", "severe")~\cite{10.1145/3442188.3445909,Yu2022.08.30.22279318}. These nuances are critical in clinical interpretation, and their omission or misclassification can lead to inaccurate scoring and overlook important differences in diagnostic meaning. Second, existing methods often provide aggregate scores without clear attribution, making it difficult to interpret results or trace errors back to specific report elements~\cite{green,doclens,ratescore}. This lack of transparency limits their usefulness in model debugging, error analysis, or human-in-the-loop validation. Third, many approaches rely either on fixed diagnostic labels, which constrain generalizability and adaptation to new conditions, or on single-step LLM-based judgments~\cite{green}. As a result, current evaluation pipelines struggle to balance clinical accuracy, interpretability, and scalability. It restricts the deployment of safe and trustworthy medical report generation.

\begin{figure*}[t]
\centerline{\includegraphics[width=2.1\columnwidth]{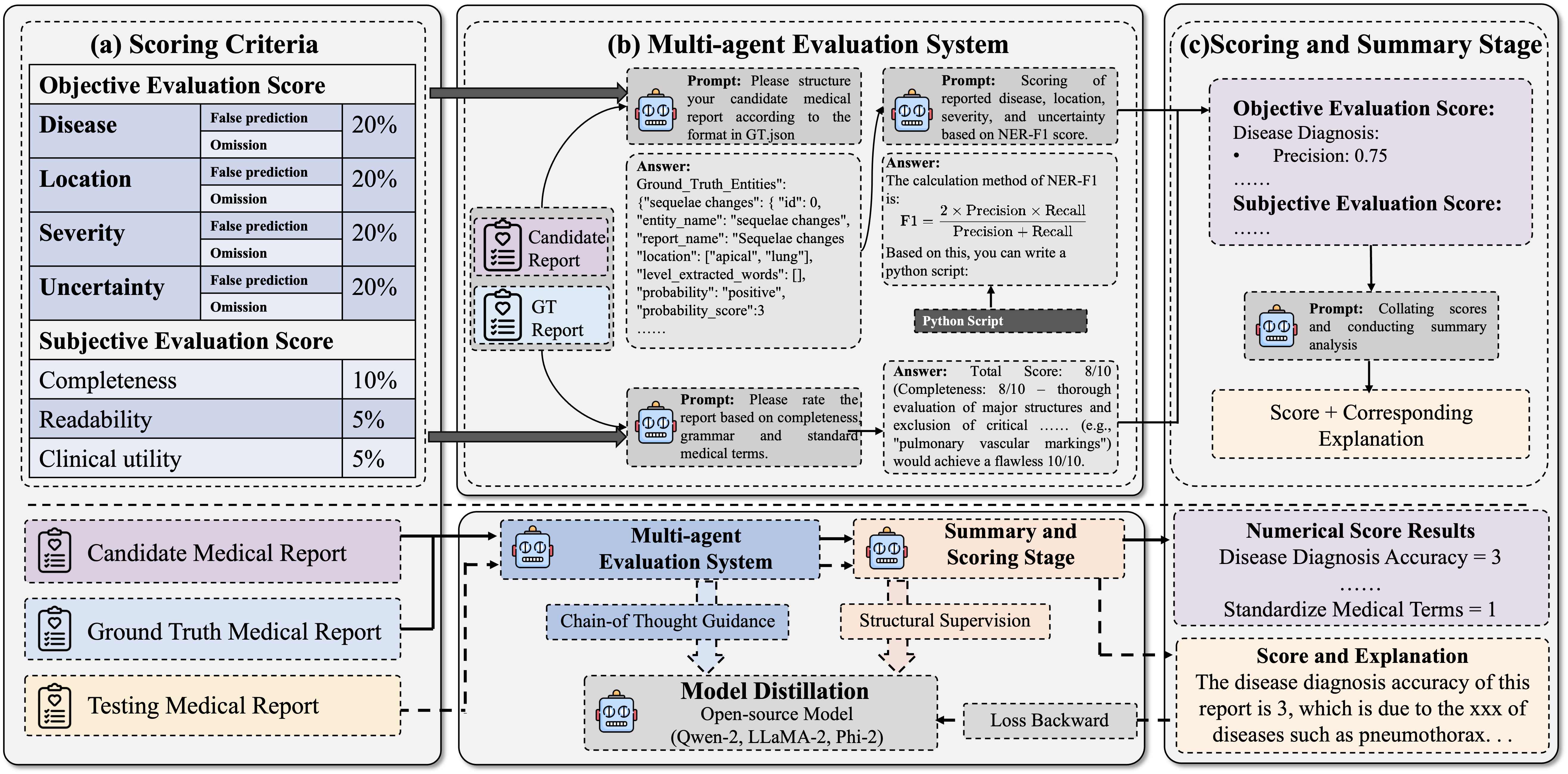}}
\caption{Workflow of our GEMA-Score. (a) The framework evaluates medical reports based on objective and subjective scoring criteria. (b) The multi-agent evaluation system assesses the candidate report against ground truth using structured prompts and automated scoring scripts. (c) The scoring and summary stage aggregates evaluation results. It provides numerical scores and detailed explanations for disease diagnosis, grammar, and terminology standardization. }
\label{fig2}
\end{figure*}

To address the limitations of existing evaluation methods and improve the explainability, clinical fidelity, and transparency of report assessment, we propose the Granular Explainable Multi-Agent Score (GEMA-Score) (Fig.~\ref{fig2}). Rather than producing a single opaque score, GEMA-Score decomposes the evaluation process into four specialized agents, each targeting a distinct aspect of report quality. This modular design directly tackles challenges by enabling fine-grained semantic understanding, interpretable error attribution, and flexible multi-criteria evaluation: (a) Entity Extraction Agent identifies key clinical findings (e.g., disease, location) from both generated and reference reports, supporting semantic alignment and synonym handling. (b) Objective Clinical Accuracy Agent computes F1 scores across four dimensions (disease, location, severity, and uncertainty) to capture diagnostic nuances. (c) Subjective Expressiveness Evaluation Agent provides a human-aligned assessment based on completeness, readability, and clinical utility, addressing aspects beyond factual correctness. (d) Score Evaluation Agent integrates the objective and subjective results to produce a comprehensive and interpretable final score. All agents operate automatically on input-output report pairs, enabling structured and transparent evaluation. GEMA-Score shows strong alignment with expert ratings (Kendall’s $\tau$ = 0.69 on ReXVal and 0.45 on RadEvalX), and improves the reliability and clinical applicability of report assessment in AMRG models. Our contributions are summarized as follows:
\begin{itemize}
    \item[$\bullet$] We construct a granular explainable multi-agent score system. It combines objective quantification and subjective evaluation.
    \item[$\bullet$] GEMA-Score generates detailed explanatory feedback to improve the verification and reliability of the report evaluation.
    \item[$\bullet$] The experimental results show that GEMA-Score is highly consistent with human expert assessment and verify its clinical application potential.
    \item[$\bullet$] We further validate the generalizability of GEMA-Score on CT report data, demonstrating its robustness across imaging modalities.
\end{itemize}

% \begin{figure}[t]
% \centerline{\includegraphics[width=\columnwidth]{fig/json_statistics.png}}
% \caption{Comparison of AI-extracted named entities vs. human accuracy and analysis of the most frequently missed words in report generation. (a) AI demonstrates high accuracy in fuzzy matching but struggles with exact matches, particularly in entity extraction. (b) The most commonly missed terms include `pleural effusion,' `lung opacity,' and `atelectasis,' with exact matches being more challenging.}
% \label{fig:entity_matching}
% \end{figure}

\begin{figure*}[t]
\centerline{\includegraphics[width=2.1\columnwidth]{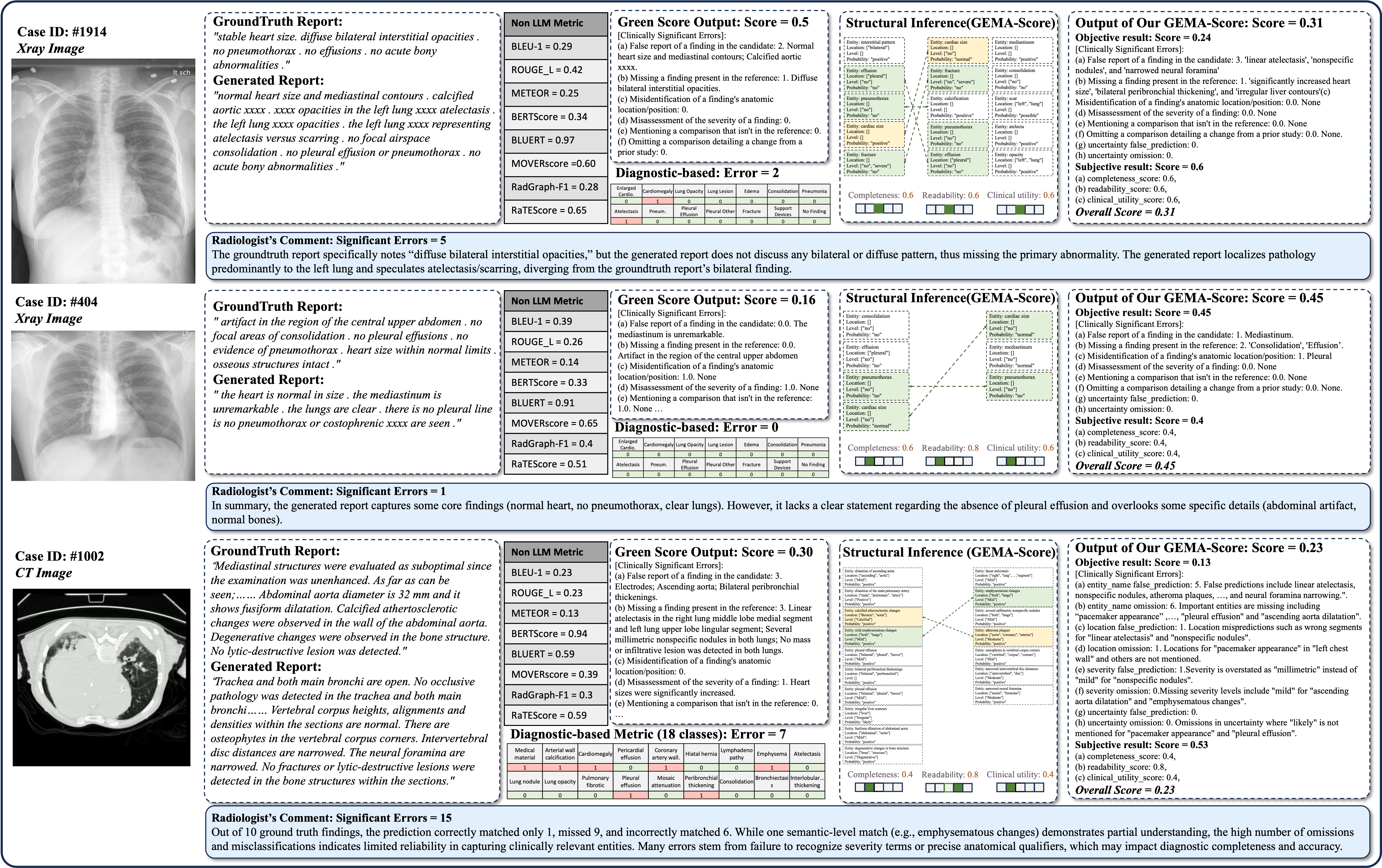}}
\caption{A multi-modal case study involving X-ray and CT images, comparing ground truth and generated reports using radiologist feedback, NLP metrics, Green-Score, and the stepwise GEMA-Score assessing clinical and linguistic quality.}
\label{fig4}
\end{figure*}

\section{Related Work}
\subsubsection{Multi-agent System.}
Multi-agent systems improve robustness and interpretability by assigning tasks to specialized agents \cite{agent1, agent3, agent4}. This has enabled dialogue coordination, tool use, and task decomposition, with agents collaborating or critiquing to enhance reliability \cite{agent4}. For instance, AutoGen \cite{autogen} uses planner, coder, and debugger agents for complex tasks, while MAD \cite{mad} improves factuality through adversarial debate. These approaches show that structured multi-agent reasoning can outperform single-agent pipelines \cite{agent3}. In clinical NLP, such collaboration holds promise for report evaluation, where distinct agents can focus on different criteria (e.g., clinical accuracy vs. language fluency), leading to more robust and interpretable assessments.

\begin{table*}[t]
\centering
\resizebox{\textwidth}{!}{\begin{tabular}{lcccc}
\toprule
\multirow{2}{*}{\textbf{Metric}} & \multicolumn{2}{c}{\textbf{Clinically Insignificant Errors}} & \multicolumn{2}{c}{\textbf{Clinically Significant Errors}} \\
 & \textbf{Kendall’s Tau$\uparrow$ (P-Value$\downarrow$)} & \textbf{Spearman$\uparrow$ (P-Value$\downarrow$)} & \textbf{Kendall’s Tau$\uparrow$ (P-Value$\downarrow$)} & \textbf{Spearman$\uparrow$ (P-Value$\downarrow$)} \\
\midrule
BLEU-1 \cite{bleu} &0.348 (8.90e-12)&0.482 (4.87e-13)&0.362 (3.15e-13)&0.513 (8.16e-15)\\
ROUGE-L \cite{rouge} &0.410 (1.19e-15)&0.558 (8.63e-18)&0.452 (1.14e-19)& 0.624 (5.19e-23)\\
METEOR \cite{meteor} &0.362 (2.03e-12)&0.500 (4.61e-14)&0.495 (5.00e-23)&0.667 (4.05e-27)\\
BertScore \cite{bertscore} &0.236 (3.66e-6)&0.339 (8.95e-07)&0.265 (1.03e-7)&0.400 (4.32e-09)\\
BLUERT \cite{bluert} &0.329 (9.55e-11)&0.461 (6.72e-12)&0.348 (2.26e-12)&0.493 (1.20e-13)\\
MOVERScore \cite{moverscore} &-0.353 (4.41e-12)&-0.490 (1.78e-13)&-0.439 (9.29e-19)&-0.603 (3.19e-21)\\
RadGraphF1 \cite{radgraph} &0.374 (7.74e-13)&0.500 (4.49e-14)&0.551  (2.17e-27)&0.717 (6.47e-33)\\
RaTEScore \cite{ratescore} &0.419 (2.87e-16)  &0.563(3.80e-18)  & 0.507(3.23e-24) & 0.682(1.17e-28) \\
Green \cite{green} &0.450 (1.07e-16)&0.596 (1.31e-20)&0.647  (1.68e-34)& 0.811 (6.61e-48)\\
\midrule
GEMA-Score (Claude-opus-4) &0.379 (1.68e-12)  & 0.498 (6.11e-14)  &  0.651 (2.25e-38) &  0.822 (2.53e-50)\\
GEMA-Score (Claude-sonnet-4) &0.361 (1.42e-11)  &0.475 (1.89e-12)  &  0.621 (3.43e-35) &  0.799 (1.56e-45) \\
GEMA-Score (Gemini-2.5-pro) &0.351 (5.89e-11)  &0.458 (8.31e-12)   &  0.643 (1.31e-37) &  0.817 (3.87e-49)\\
GEMA-Score (Deepseek-v3)& 0.381 (1.36e-12) & 0.500 (4.93e-14)  & 0.661 (1.45e-39) & 0.832 (1.81e-52) \\
GEMA-Score (Deepseek-r1) &0.361 (2.39e-11) & 0.468 (2.69e-12)  & 0.639 (2.84e-37) & 0.816 (5.66e-49)\\
GEMA-Score (Chat-GPT-4o) &  0.367 (8.07e-12) & 0.480 (6.21e-13) & 0.632 (2.47e-36) & 0.810 (9.98e-48) \\
GEMA-Score (Chat-GPT-o1) &0.372 (3.49e-12)  & 0.477 (9.26e-13) &  0.627 (8.67e-36) & 0.803 (2.28e-46) \\
GEMA-Score (Chat-GPT-o3) &0.371 (4.48e-12) & 0.389 (1.95e-13)&  0.672 (8.24e-41) & 0.846 (6.58e-56) \\
\midrule
\textbf{GEMA-Score (Distilled LLaMA-3.1-8B)} & \textbf{0.465 (2.36e-17)} & \textbf{0.586 (8.14e-20)} & \textbf{0.678 (9.16e-41)} & \textbf{0.845 (8.48e-56)} \\
\bottomrule
\end{tabular}}
\caption{Clinical Significance: Human Correlation Comparison of Evaluation Metrics on ReXVal Dataset}
\label{tab:ReXVal_comparison}
\end{table*}

\begin{figure*}[t]
\centerline{\includegraphics[width=2.1\columnwidth]{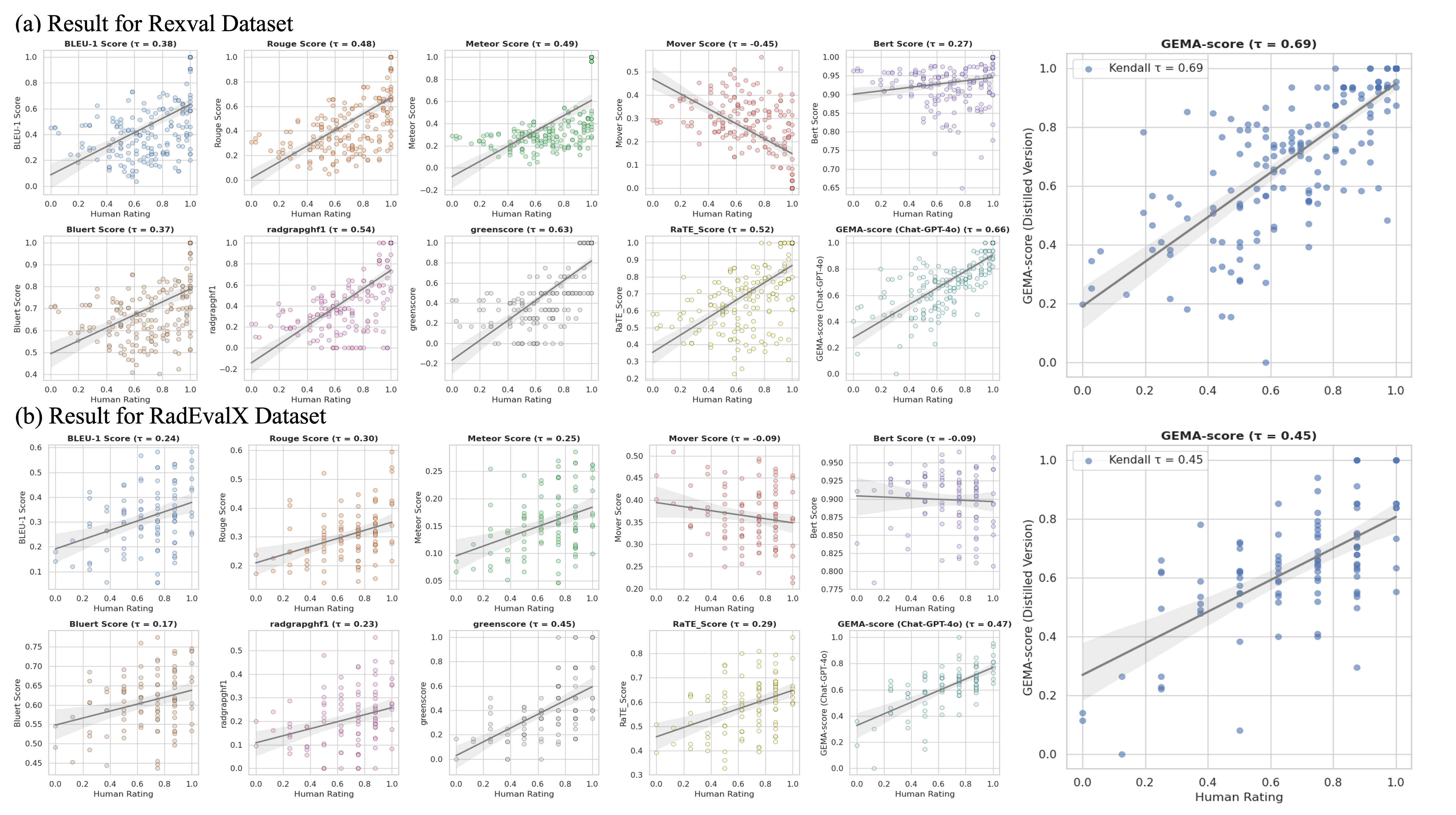}}
\caption{Correlation coefficients with radiologists. (a) Comparison against other metrics on the ReXVal dataset. (b) Comparison against other metrics on the RadEvalX dataset. The rightmost panel is the result of distilled GEMA-score(LLaMA-3.1-8B).}
\label{fig5}
\end{figure*}
%\begin{figure*}[t]\centerline{\includegraphics[width=\columnwidth]{fig/granular_score.png}}\caption{Results in RadEvalX Benchmark: Correlation Coefficients with Radiologists Results }\label{fig3}\end{figure*}
\begin{table*}[t]
\centering
\resizebox{\textwidth}{!}{\begin{tabular}{lcccc}
\toprule
\multirow{2}{*}{\textbf{Metric}} & \multicolumn{2}{c}{\textbf{Clinically Insignificant Errors}} & \multicolumn{2}{c}{\textbf{Clinically Significant Errors}} \\
 & \textbf{Kendall’s Tau↑ (P-Value↓)} & \textbf{Spearman↑ (P-Value↓)} & \textbf{Kendall’s Tau↑ (P-Value↓)} & \textbf{Spearman↑ (P-Value↓)} \\
\midrule
BLEU-1 \cite{bleu} &0.122 (0.111)&0.160 (0.112)&0.147 (0.048)&0.195 (0.052)\\
ROUGE-L \cite{rouge} &0.104 (0.175)&0.130 (0.197)&0.201 (0.007)&0.259 (0.009)\\
METEOR \cite{meteor} &0.114 (0.134)&0.148 (0.143)&0.154 (0.039)&0.208 (0.038)\\
BertScore \cite{bertscore} &-0.024 (0.754)&-0.029 (0.777)&-0.101 (0.175)&-0.127 (0.207)\\
BLUERT \cite{bluert} &0.100 (0.191)&0.134 (0.183)&0.090 (0.225)&0.126 (0.212)\\
MOVERScore \cite{moverscore} &-0.084 (0.272)&-0.111 (0.271)&-0.005 (0.946)&-0.005 (0.963)\\
RadGraphF1 \cite{radgraph} &0.135 (0.078)&0.179 (0.075)&0.130 (0.082)&0.183 (0.069)\\
RaTEScore \cite{ratescore} &0.165 (0.031)  &0.214 (0.032)  &0.177 (0.018)  &0.238(0.017)  \\
Green \cite{green} &0.118 (0.134)&0.141 (0.163)&0.347 (6.46e-6)&0.433 (6.99e-6)\\
\midrule
GEMA-Score (Claude-opus-4) &  0.143 (0.075) & 0.175 (0.081)  & 0.423 (1.38e-08) & 0.551(2.86e-09) \\
GEMA-Score (Claude-sonnet-4) &  0.173 (0.032) & 0.216 (0.031)  &  0.416 (2.50e-08) & 0.542(5.71e-09) \\
GEMA-Score (Gemini-2.5-pro)&  0.154 (0.056) & 0.190 (0.059)    & 0.399 (8.94e-08) & 0.521(2.71e-08)\\
GEMA-Score (Deepseek-v3) &  0.173 (0.032) & 0.209 (0.037)  & 0.465 (4.39e-10) & 0.608(1.95e-11) \\
GEMA-Score (Deepseek-r1) &  0.185 (0.022) & 0.226 (0.024)  & 0.424 (1.28e-08) & 0.549(3.21e-09) \\
GEMA-Score (Chat-GPT-4o) &  0.158 (0.050) & 0.190 (0.057)  & 0.452 (1.31e-09) & 0.573(4.59e-10) \\
GEMA-Score (Chat-GPT-o1) &  0.185 (0.021) & 0.229 (0.021)  & 0.466 (3.98e-10) & 0.597(5.50e-11) \\
GEMA-Score (Chat-GPT-o3) &  0.176 (0.029) & 0.219 (0.029)  & 0.444 (2.51e-09) & 0.573(4.77e-10)\\
\midrule
\textbf{GEMA-Score (Distilled LLaMA-3.1-8B)} & \textbf{0.133 (0.102)} & \textbf{0.167 (0.097)} & \textbf{0.414 (3.07e-8)} & \textbf{0.533 (1.15e-8)} \\
\bottomrule
\end{tabular}}
\caption{Clinical Significance: Human Correlation Comparison of Evaluation Metrics on RadEvalX Dataset}
\label{tab:RadEvalX_comparison}
\end{table*}

\subsubsection{Medical Report Evaluation Metric.}

Various automatic metrics have been proposed to evaluate the generated medical reports. Early metrics like BLEU \cite{bleu}, ROUGE \cite{rouge}, and METEOR \cite{meteor} rely on surface-level n-gram overlap and struggle with lexical variation, especially in clinical contexts. Semantic similarity metrics such as BERTScore \cite{bertscore}, MOVERScore \cite{moverscore}, and BLUERT \cite{bluert} leverage contextual embeddings to capture meaning, but often conflate fluency with factual accuracy and inflate scores for redundant or vague content. Structure-aware metrics such as RadGraphF1 \cite{radgraph} offer greater clinical relevance by extracting clinical entities and relations; however, they depend on accurate entity extraction and may miss subtle semantic errors. More recently, LLM-based evaluators aim to approximate expert judgment. GREEN \cite{green} uses free-form LLM rationales; RaTEScore \cite{ratescore} and DocLens \cite{doclens} incorporate aspect-specific prompts; and CheXagent \cite{chexagent} applies condition-focused LLM agents. Although promising, these methods still face challenges in explainable and fine-grained attribution. 

\section{Method}
\subsubsection{Problem Definition and Scoring Framework.}
Automatic medical report generation has the potential to improve the traditional cumbersome medical process. However, a comprehensive evaluation of these report generation models remains challenging due to the dual nature of clinical assessment. On the one hand, an effective generated report must maintain objective clinical accuracy, ensuring that diseases and findings are correctly characterized. On the other hand, the report should exhibit high linguistic quality, aligning with expert radiologists’ reporting styles and terminology preferences. To address these challenges, we formalize a composite scoring task with two-fold objectives. Objective evaluation $S_{obj}$ quantifies accuracy in disease characterization based on four clinical granularities (disease entity, location, severity, uncertainty). Subjective evaluation $S_{sub}$ assesses the completeness, readability, and clinical usability of the diagnostic report. The overall score $S_{overall}$ can be defined as:
\begin{equation}
    S_{overall} = \sum_{i=1}^{K} w^{(i)}*S^{(i)}_{obj}+\sum_{i=1}^{I} w^{(i)}*S^{(i)}_{sub},
\end{equation}
where K and I represent categories defined for objective and subjective evaluations.
This framework integrates both clinical precision and linguistic quality. It provides a holistic evaluation strategy for medical AI systems and bridges the gap between automated assessment and real-world radiological reporting standards.

\noindent
\textbf{Entity Extraction Agent.} 
Given a generated report $\mathcal{R}$, the entity extraction agent segments it into fine-grained entities, including \textit{disease}, \textit{location}, \textit{severity}, and \textit{uncertainty}. The structured output is represented as:
% \begin{equation}
% \mathit{E} = \{ (\mathit{E}_d^{(i)}, \mathit{E}_l^{(i)}, \mathit{E}_s^{(i)}, \mathit{E}_u^{(i)}) \}_{i=1}^{N}
% \end{equation}
% where $\mathit{E}_d^{(i)}$ denotes the disease, $\mathit{E}_l^{(i)}$ represents the associated location, $\mathit{E}_s^{(i)}$ indicates severity, and $\mathit{E}_u^{(i)}$ captures uncertainty.
\begin{equation}
\label{eq:entities}
\mathit{E} = \Bigl\{ 
\bigl(\mathit{E}_d^{(i)}, \mathit{E}_l^{(i)}, \mathit{E}_s^{(i)}, \mathit{E}_u^{(i)}\bigr) 
\Bigr\}_{i=1}^{N},
\end{equation}
where \(\mathit{E}_d^{(i)}\) denotes the disease entity, \(\mathit{E}_l^{(i)}\) represents the corresponding anatomical location, \(\mathit{E}_s^{(i)}\) indicates severity, and \(\mathit{E}_u^{(i)}\) captures uncertainty descriptors. This structured representation enables granular evaluation of radiology reports.

\subsubsection{Objective Clinical Accuracy Agent.}
To quantitatively assess the factual consistency between the generated report $\hat{x}$ and the reference report $x$, we define a structured matching protocol over clinically significant entity types. Specifically, each report is decomposed into entity tuples spanning four semantic dimensions: disease, location, severity, and uncertainty. These entity sets are denoted as $E(x)$ and $E(\hat{x})$.

For each dimension, the agent performs type-specific comparisons between matched entity sets. The similarity score from reference to generation is defined as:
\begin{equation}
S(x, \hat{x}) = \frac{1}{|E(\hat{x})|} \sum_{e_j \in E(\hat{x})} \mathbb{I}\left( \exists e_i \in E(x) : \text{match}(e_j, e_i) \right)
\end{equation}
where $\mathbb{I}$ is a binary condition indicator function. A symmetric comparison $S(\hat{x}, x)$ is also computed.

To reflect both precision and recall across entity types, we define the objective clinical accuracy score as the harmonic mean:
\begin{equation}
    \text{S}_{\text{obj}} = 
    \begin{cases}
    0, & \text{if } S(x, \hat{x}) + S(\hat{x}, x) = 0 \\
    \frac{2 \cdot S(x, \hat{x}) \cdot S(\hat{x}, x)}{S(x, \hat{x}) + S(\hat{x}, x)}, & \text{otherwise}
    \end{cases}
\end{equation}
Each sub-dimension contributes independently and may optionally be weighted to reflect clinical importance.

\subsubsection{Subjective Expressiveness Evaluation Agent.}

To evaluate the linguistic quality of the generated report, we introduce a subjective expressiveness score based on fluency, grammar, and medical terminology usage. Let $\mathcal{E} = \{\text{fluency}, \text{grammar}, \text{terminology}\}$ denote the evaluated dimensions. For each aspect $a \in \mathcal{E}$, the agent identifies binary error indicators $\text{err}_{a}^{(k)}$ over observed issues.

% The raw score for each dimension is defined as:
% \begin{equation}
%     \text{S}_a = 1 - \sum_k \text{err}_a^{(k)},
% \end{equation}
The final subjective score is computed as a weighted sum:
% \begin{equation}
%     \text{S}_{\text{sub}} = \sum_{a \in \mathcal{E}} w_a \cdot \frac{1}{1 + \sum \text{err}_a^{(k)}}
% ,
% \end{equation}
\begin{equation}
    \text{S}_{\text{sub}} = \sum_{a \in \mathcal{E}} w_a \cdot \max\left(0, 1 - \lambda \cdot \sum \text{err}_a^{(k)} \right),
\end{equation}

where $w_a$ is the agent-defined importance weight and $\lambda$ is penalty strength and is set to 0.05. To ensure interpretability, $\text{S}_{\text{sub}}$ is rounded to the nearest value in the discrete set $\{0.0, 0.2, 0.4, 0.6, 0.8, 1.0\}$.

\subsubsection{Score Evaluation Agent.}

The final GEMA-Score aggregates factual and linguistic assessments to yield a comprehensive evaluation of the generated report:
\begin{equation}
    \text{GEMA-Score}(x, \hat{x}) = \alpha \cdot \text{S}_{\text{obj}} + (1 - \alpha) \cdot \text{S}_{\text{sub}},
\end{equation}
where $\alpha \in [0,1]$ balances clinical accuracy and linguistic expressiveness. We set $\alpha = 0.8$ by default to emphasize factual correctness.

Unlike prior metrics that output only scalar scores, our score evaluation agent additionally provides structured outputs following a predefined format, including entity-level false predictions, omissions, and detailed textual explanations for each clinical aspect (e.g., location, severity, uncertainty). This enhances the transparency and interpretability of the overall evaluation process.
% \multirow{6}{*}{\textbf{AMRG Model}} 
% & R2GEN \cite{r2gen}       & 0.326 & 0.276 & 0.159 & 0.933 & 0.630 & 0.340 & 0.137 & 0.276 & 0.534 \\
% & R2GEN-CMN \cite{r2gencmn} & 0.340 & 0.284 & 0.168 & 0.934 & 0.637 & 0.334 & 0.148 & 0.295 & 0.530 \\
% & RGRG \cite{rgrg}         & 0.168 & 0.146 & 0.158 & 0.907 & 0.538 & 0.370 & 0.152 & 0.246 & 0.553 \\
% & CheXagent \cite{chexagent} & 0.155 & 0.163 & 0.186 & 0.904 & 0.529 & 0.359 & 0.165 & 0.263 & 0.537 \\
% & RadFM \cite{radfm}       & 0.067 & 0.081 & 0.049 & 0.787 & 0.427 & 0.535 & 0.001 & 0.000 & 0.201 \\
% & Mini-GPT \cite{minigpt-med} & 0.054 & 0.078 & 0.031 & 0.753 & 0.404 & 0.577 & 0.012 & 0.007 & 0.233 \\\hline
% \multirow{3}{*}{\makecell{\textbf{Open-source}\\\textbf{Model}}} 
% & InternVL2 \cite{chen2024far}             & 0.040 & 0.085 & 0.043 & 0.887 & 0.464 & 0.455 & 0.062 & 0.119 & 0.487 \\ 
% & LLAVA \cite{NEURIPS2023_6dcf277e}                 & 0.070 & 0.084 & 0.043 & 0.784 & 0.426 & 0.534 & 0.016 & 0.024 & 0.280 \\
% & HuatuoGPT-o1 \cite{chen2024huatuogpt}          & 0.081 & 0.089 & 0.057 & 0.851 & 0.466 & 0.497 & 0.062 & 0.089 & 0.451 \\

\begin{table*}[t]
\centering
\resizebox{\textwidth}{!}{%
\begin{tabular}{l|ccc|ccc|c|ccc|c}
% \hline
%\multicolumn{9}{l}{}\\
% \hline
% & \multicolumn{3}{c|}{\textbf{Overlap-Based}} & \multicolumn{3}{c|}{\textbf{BERT-Based}} & \multicolumn{1}{c|}{\textbf{NER-F1}}& \multicolumn{1}{c}{\textbf{LLM-based}}\\
% \cline{2-9}
% \textbf{Analysis Method}  & 
% \makecell{BLEU-1 \\ \cite{bleu}} & 
% \makecell{ROUGE \\ \cite{rouge}} & 
% \makecell{METEOR \\ \cite{meteor}} & 
% \makecell{BERTScore \\ \cite{bertscore}} & 
% \makecell{BLUERT \\ \cite{bluert}} & 
% \makecell{MOVERScore \\ \cite{moverscore}} & 
% \makecell{RadGraph-F1 \\ \cite{radgraph}} & 
% \makecell{Green \\ \cite{green}} \\ 
% \hline
% kendall Correlation       &0.189               & 0.177              & 0.200              & 0.137            &0.211               &  -0.183               & 0.054              & 0.326              \\
% Significance (p-value)    & 0.006              & 0.011              &0.004               & 0.050              & 0.002              &0.008               & 0.436              &0.000               \\ 
% \hline
% \multicolumn{9}{l}{\textbf{(a) Statistics between our GEMA-Score and Existing Evaluation Score in X-ray:}}\\
\hline
& \multicolumn{3}{c|}{\textbf{Overlap-Based}} & 
  \multicolumn{3}{c|}{\textbf{BERT-Based}} & 
  \textbf{NER-F1} & 
  \multicolumn{3}{|c|}{\textbf{Diagnostic-Based}} & 
  \textbf{LLM-based} \\
\cline{2-12}
\textbf{Analysis Method} & BLEU-1 & ROUGE & METEOR & BERTScore & BLUERT & MOVERScore & RadGraph-F1& Chexbert$_P$ & Chexbert$_R$ &Chexbert$_{F1}$ & Green \\
\hline
kendall Correlation    & 0.189 & 0.177 & 0.200 & 0.137 & 0.211 & -0.183 & 0.054 & 0.025 & 0.009 & 0.021 & 0.326 \\
Significance (p-value) & 0.006 & 0.011 & 0.004 & 0.050 & 0.002 & 0.008 & 0.436 & 8.450e-01 & 9.421e-01 & 8.7002e-01 & 0.001 \\
\hline
\end{tabular}}
\caption{Correlation and significance between GEMA-Score predictions and different medical report evaluation metric.}
\label{tab:metrics}
\end{table*}

\section{Experiment and Analysis}
% \subsection{Experiment Setting}
\subsubsection{Dataset and Evaluation Settings.} 
In this study, we conduct experiments on five datasets: MIMIC-CXR~\cite{Johnson2019}, ReXVal~\cite{Yu2022.08.30.22279318,Yu2023,Goldberger2000}, RadEvalX~\cite{Calamida2024,calamida2023radiologyawaremodelbasedevaluationmetric,Goldberger2000}, and CT-RATE~\cite{hamamci2024developing}. MIMIC-CXR includes over 377,000 chest X-rays and 227,000 reports, from which 3,000 cases are selected for metric correlation analysis. ReXVal contains 50 studies annotated by six radiologists, identifying clinically significant and insignificant errors across six categories. RadEvalX includes 100 studies with similar annotations from two radiologists, covering eight error types. Both datasets are used to assess alignment between automatic scores and expert evaluations. To evaluate the generalizability of GEMA-Score beyond X-rays, we further test it on chest CT data using the CT-RATE dataset. We randomly selected 60 studies with paired reference and generated reports, and conducted human expert annotation of clinical errors with a focus on entity names and locations. 

The LLM-based agents operate under deterministic decoding settings, with temperature set to 0 and top-p set to 1, ensuring consistent and reproducible outputs. A maximum token limit of 8192 is used to handle long-form clinical inputs and structured outputs.

\subsection{Experimental Results.}

% \subsubsection{Named Entity Matching Evaluation.}
% Fig.~\ref{fig:entity_matching}(a) compares the accuracy of AI-extracted named entities to human-labeled references compared to entities in the cad-chest dataset. In particular, our model demonstrates high accuracy when applying fuzzy matching to pair similar or morphologically varied terms (over $90\%$ for most terms). Fig.~\ref{fig:entity_matching}(b) shows the most commonly missed words, indicating that the fuzzy matching approach effectively captures semantically related terms, improving entity extraction performance in clinically relevant but variably expressed findings.
\subsubsection{Case Study Analysis.}
Fig.~\ref{fig4} presents three representative case studies to demonstrate how GEMA-Score offers clinically aligned evaluations, especially in both X-ray and CT modalities. In Case $\#$1914 (X-ray), the generated report omits ``diffuse bilateral interstitial opacities'', a clinically significant finding; yet traditional metrics (BLEU, METEOR, RadGraph-F1) assign moderate scores ($0.25$–$0.5$). In contrast, GEMA-Score assigns a lower score ($0.31$), consistent with expert annotations highlighting the omission as a major diagnostic error. Case $\#$404 (X-ray) contains only minor omissions; GEMA-Score assigns a higher score ($0.45$), consistent with the expert-rated Error = 1. In Case $\#$1002 (CT), the generated report misses multiple findings related to vertebral changes and degenerative disease. Despite partial semantic matches, diagnostic-based methods yield misleadingly lenient errors $= 7$. GEMA-Score instead penalizes the omissions and misclassifications with a lower score ($0.23$), better reflecting the expert-rated 15 significant errors. Overall, GEMA-Score demonstrates stronger alignment with radiologist assessments. Detailed scores and additional examples are provided in the supplementary materials.

\subsubsection{Benchmarking Metrics Against Human Expert Judgments.}
Tables~\ref{tab:ReXVal_comparison} and~\ref{tab:RadEvalX_comparison} compare the human correlation of different evaluation metrics in identifying clinically significant errors. GEMA-Score consistently outperforms existing metrics across both datasets. While the LLM-based Green score achieves strong correlation with expert ratings (e.g., Spearman = 0.816), GEMA-Score achieves even higher agreement (Spearman = 0.845). Notably, we benchmarked GEMA-Score across a wide range of LLM backbones and APIs, including Claude, Gemini, Chat-GPT, and Deepseek (Kendall’s $\tau$: 0.622–0.678 on ReXVal). In addition, we introduce a distilled version of GEMA-Score based on LLaMA-3.1-8B, which achieves satisfactory correlation with human experts on both datasets (e.g., Spearman: 0.845 / 0.533). This demonstrates that GEMA-Score remains robust and effective across model choices while supporting open-source deployment. Fig~\ref{fig5} our score performs the best in correlation analysis across both datasets when combining both error types. Table~\ref{tab:metrics} shows the correlation between GEMA-Score and existing metrics. Among them, Green shows the highest correlation ($\tau = 0.326$), yet a significant difference remains ($p = 0.001$), suggesting that GEMA-Score captures distinct aspects aligned with expert evaluations.

\begin{table}[t]
\centering
\resizebox{\columnwidth}{!}{
\begin{tabular}{lcccccccc}
\toprule
       & 0 & 1 & 2 & 3 & 4 & 5 & GEMA\textsuperscript{\dag} & GEMA\textsuperscript{\ddag} \\
\midrule
0     & --    & 0.554 & 0.737 & 0.703 & 0.742 & 0.617 & 0.609 &  0.636      \\
1     & 0.554 & --    & 0.591 & 0.603 & 0.681 & 0.517 & 0.457 &  0.495       \\
2     & 0.737 & 0.591 & --    & 0.745 & 0.730 & 0.793 & 0.621 &  0.677       \\
3     & 0.703 & 0.603 & 0.745 & --    & 0.747 & 0.644 & 0.640 &    0.670    \\
4     & 0.742 & 0.681 & 0.730 & 0.747 & --    & 0.612 & 0.655 &   0.725     \\
5     & 0.617 & 0.517 & 0.793 & 0.644 & 0.612 & --    & 0.530 &    0.582      \\
GEMA\textsuperscript{\dag}  & 0.609 & 0.457 & 0.621 & 0.640 & 0.655 & 0.530 & --    &    0.776      \\
GEMA\textsuperscript{\ddag} & 0.636      &   0.495  &0.677  & 0.670      &  0.725     &   0.582          &  0.776     & --     \\
\bottomrule
\end{tabular}
}
\vspace{0.5em}
\begin{minipage}{\columnwidth}
\tiny
\textsuperscript{\dag}~GEMA-Score (Chat-GPT-4o) \quad
\textsuperscript{\ddag}~GEMA-Score (Distilled LLaMA-3.1)
\end{minipage}
\caption{Pearson correlation between GEMA-Score predictions and ReXVal expert annotations across different expert raters (0–5). Higher values indicate stronger agreement.}
\label{tab:correlation}
\end{table}

\begin{table}[t]
\centering
\resizebox{\columnwidth}{!}{
\begin{tabular}{lcccccc}
\toprule
\textbf{Rater} & (a) & (b) & (c) & (d)  \\
\midrule
Expert 0 & 0.468 (4.00e-23) & 0.470 (2.47e-23) & 0.461 (1.76e-22) & 0.462 (1.39e-22)  \\
Expert 1 & 0.526 (7.99e-30) & 0.520 (4.41e-29) & 0.424 (7.13e-19) & 0.387 (1.06e-15)\\
Expert 2 & 0.600 (2.03e-40) & 0.582 (1.36e-37) & 0.539 (1.52e-31) & 0.561 (1.49e-34)  \\
Expert 3 & 0.645 (1.70e-48) & 0.590 (6.84e-39) & 0.550 (5.24e-33) & 0.514 (2.30e-28) \\
Expert 4 & 0.529 (3.17e-30) & 0.603 (6.90e-41) & 0.506 (2.30e-27) & 0.415 (4.68e-18) \\
Expert 5 & 0.563 (9.16e-35) & 0.497 (2.68e-26) & 0.395 (2.11e-16) & 0.426 (4.75e-19) \\
\midrule
GEMA\textsuperscript{\dag} & 0.591 (5.58e-19)&  0.487 (1.58e-12)&  0.507 (1.83e-14) &  0.563 (4.08e-18)   \\
GEMA\textsuperscript{\ddag} & 0.651 (2.29e-25) & 0.512 (1.15e-14)& 0.520 (2.97e-15)& 0.503 (3.22e-14)\\
\bottomrule
\end{tabular}
}
\vspace{0.5em}
\begin{minipage}{\columnwidth}
\tiny
\textsuperscript{\dag}~GEMA-Score (Chat-GPT-o1) \quad
\textsuperscript{\ddag}~GEMA-Score (Distilled LLaMA-3.1)
\end{minipage}
\caption{Pearson correlation between GEMA-Score predictions and ReXVal expert to the mean rater. Error types: 
(a) false positive finding, (b) missed reference finding, (c) misidentified location, (d) incorrect severity.}
\label{tab:finegrained}
\end{table}

\begin{table}[t]
\centering
\resizebox{\columnwidth}{!}{
\begin{tabular}{lcccc}
\toprule
\textbf{Distilled Model} &
\multicolumn{2}{c}{\textbf{RexVal}} &
\multicolumn{2}{c}{\textbf{RadEvalX}} \\
\cmidrule(lr){2-3}\cmidrule(lr){4-5}
& Sig. Corr. $\uparrow$ & Insig. Corr. $\uparrow$ & Sig. Corr. $\uparrow$ & Insig. Corr. $\uparrow$ \\
\midrule
\makecell{LLaMA-3.1-8B \\ \cite{LLaMA}}     
    & \makecell{0.678 \\ (9.16e-41)} 
    & \makecell{0.465 \\ (2.36e-17)} 
    & \makecell{0.414 \\ (3.07e-8)} 
    & \makecell{0.133 \\ (0.102)} \\
\makecell{Phi-2-2.7B \\ \cite{Phi}}     
    & \makecell{0.676 \\ (1.90e-40)} 
    & \makecell{0.459 \\ (1.89e-16)} 
    & \makecell{0.415 \\ (6.67e-9)} 
    & \makecell{0.131 \\ (0.024)} \\
\makecell{Qwen-2VL-2B \\ \cite{Qwen}}       
    & \makecell{0.648 \\ (4.86e-37)} 
    & \makecell{0.422 \\ (4.60e-14)} 
    & \makecell{0.369 \\ (8.94e-7)} 
    & \makecell{0.221 \\ (0.008)} \\
\bottomrule
\end{tabular}}
\caption{Correlation on significant and insignificant findings across RexVal and RadEvalX with different distillation model.}
\label{tab:distil}
\end{table}

\begin{table}[ht]
\centering
\resizebox{\columnwidth}{!}{
\begin{tabular}{lccc}
\toprule
\textbf{Agent Setting} & \textbf{Struct. Error} & \multicolumn{2}{c}{\textbf{F1 Score}} \\
\cmidrule(lr){3-4}
& mean (std) $\downarrow$& Kendall (p) $\uparrow$ & Spearman (p) $\uparrow$ \\
\midrule
BLEU-1 \cite{bleu}& -- & 0.231 (2.3e-02) & 0.291 (2.4e-02)\\
ROUGE-L \cite{rouge} & -- & 0.218 (3.1e-02) & 0.281 (3.0e-02 )\\
BLUERT \cite{bluert}& -- & 0.226 (2.5e-02) & 0.285 (2.7e-02)\\
RadGraphF1 \cite{radgraph} & -- & 0.224 (2.7e-02) & 0.288 (2.6e-02)\\
Radbert~\cite{radbert}& -- & 0.170 (1.0e-01) & 0.209 (1.1e-01)\\
GREEN \cite{green} & -- & 0.310 (2.6e-03) & 0.381 (2.7e-03) \\
\midrule
GEMA-Score (Deepseek-V3) & 0.900 (1.674) & 0.576 (1.7e-07) & 0.632 (7.9e-08) \\
GEMA-Score (Deepseek-R1) & 1.406 (2.275) & 0.482 (3.8e-06) & 0.526 (1.8e-05) \\
GEMA-Score (Chat-GPT-4o) & 0.750 (1.451) & 0.592 (2.8e-08) & 0.668 (5.2e-09) \\
GEMA-Score (Chat-GPT-o1) & 0.733 (1.191) & 0.755 (9.7e-13) & 0.794 (3.8e-14) \\
GEMA-Score (Chat-GPT-o3) & 0.850 (1.459) & 0.712 (1.6e-11) & 0.774 (4.0e-13) \\
\midrule
GEMA-Score (LLaMA-3.1-8B) & -- & 0.660 (1.0e-09) & 0.738 (1.7e-11) \\
GEMA-Score (Phi-2-2.7B) & -- & 0.341 (1.9e-03) & 0.377 (2.9e-03) \\
GEMA-Score (Qwen-2VL-2B) & -- & 0.628 (4.4e-09) & 0.708 (2.4e-10) \\
\midrule
GEMA-Score (LLaMA-3.1$\&$CoT) & 0.633 (1.301) & 0.689 (5.4e-11) & 0.741 (1.2e-11) \\
\bottomrule
\end{tabular}
}
\caption{Assessment of Structured Extraction and F1 Score Consistency on CT Reports.}
\label{tab:ct_eval}
\end{table}

% \begin{table}[t]
% \centering
% \caption{Difference between ReXVal experts and the GEMA-score model measured using mean absolute error of significant error counts.}
% \resizebox{\columnwidth}{!}{
% \begin{tabular}{lccccccc}
% \toprule
%       & 0     & 1     & 2     & 3     & 4     & 5     & GEMA-score \\
% \midrule
% 0     & --    & 0.505 & 0.835 & 0.675 & 0.495 & 1.130 & 1.160 \\
% 1     & 0.505 & --    & 1.100 & 0.870 & 0.660 & 1.365 & 1.485 \\
% 2     & 0.835 & 1.100 & --    & 0.730 & 0.770 & 0.725 & 0.715 \\
% 3     & 0.675 & 0.870 & 0.730 & --    & 0.570 & 0.965 & 0.895 \\
% 4     & 0.495 & 0.660 & 0.770 & 0.570 & --    & 1.025 & 1.005 \\
% 5     & 1.130 & 1.365 & 0.725 & 0.965 & 1.025 & --    & 0.930 \\
% GEMA-score & 1.160 & 1.485 & 0.715 & 0.895 & 1.005 & 0.930 & --    \\
% \bottomrule
% \end{tabular}}
% \end{table}

\subsection{Analysis and Discussion.}
\subsubsection{Consistency Analysis for Expert-Level Evaluation.}
Table~\ref{tab:correlation} reports Pearson correlations between GEMA-Score predictions and ReXVal expert annotations on clinically significant error counts. GEMA$^\dagger$ (Chat-GPT-4o) achieves an average correlation of 0.585, approaching the average inter-expert agreement (0.668). The distilled version, GEMA$^\ddagger$ (LLaMA-3.1), yields an even higher consistency with a mean correlation of 0.630, demonstrating its reliability in a lightweight setting. Table~\ref{tab:finegrained} further analyzes model behavior across four error types. The distilled GEMA$^\ddagger$ achieves the highest average correlation with experts across all categories, reaching 0.651 for false positives (a), 0.512 for missed findings (b), 0.520 for location errors (c), and 0.503 for severity errors (d). These results indicate strong alignment with expert fine-grained judgments. Further, the distilled GEMA$^\ddagger$ yields consistently higher agreement, validating its reliability for fine-grained clinical assessment.

\subsubsection{Distilled Model Evaluation for Local Deployment.}
Table~\ref{tab:distil} shows that LLaMA-3.1-8B achieves the best overall performance, with the highest correlation on ReXVal for both significant (0.678) and insignificant errors (0.465), and strong performance on RadEvalX (0.414 / 0.133). Phi-2-2.7B offers comparable results (0.676 / 0.459 on ReXVal and 0.415 / 0.131 on RadEvalX), making it a practical alternative for resource-constrained environments. Qwen-2VL-2B lags behind with lower correlation values, especially for ReXVal significant errors (0.648) and RadEvalX significant errors (0.369). 
% Overall, LLaMA-3.1-8B provides the most robust and balanced distilled option for local deployment.

\subsubsection{Generalization Analysis for CT Report Evaluation.}
Table~\ref{tab:ct_eval} presents GEMA-Score performance on CT reports from the CT-RATE dataset, demonstrating its generalizability. GEMA-Score consistently outperforms the single-step baseline GREEN across all metrics (e.g., Kendall correlation from 0.310 to 0.755). The best results are achieved by Chat-GPT-o1 (Struct. Error: 0.733$\pm$1.191), with strong performance also observed from Chat-GPT-o3 and Chat-GPT-4o. Among distilled models, LLaMA-3.1-8B performs best (Kendall: 0.660, Spearman: 0.738), and the LLaMA-3.1+Chain-of-Thought variant achieves further improvements (Kendall: 0.689, Spearman: 0.741) while yielding the lowest structural error (0.633$\pm$1.301). These results underscore the value of multi-agent, multi-step reasoning for robust clinical report evaluation across imaging modalities.

\section{Conclusion}
GEMA-Score offers a structured multi-agent framework that assesses both objective clinical accuracy and subjective report quality. It aligns well with expert judgments, validating its clinical reliability. Further, we introduce a distilled version for local deployment and extend its use from chest X-rays to CT reports, demonstrating cross-modality generalizability. These advances strengthen GEMA-Score as a scalable and trustworthy tool for medical report evaluation.

\section{Acknowledgements}
Guang Yang was supported in part by the ERC IMI (101005122), the H2020 (952172), the MRC (MC/PC/21013), the Royal Society (IEC/NSFC/211235), the NVIDIA Academic Hardware Grant Program, the SABER project supported by Boehringer Ingelheim Ltd, NIHR Imperial Biomedical Research Centre (RDA01), The Wellcome Leap Dynamic resilience program (co-funded by Temasek Trust), UKRI guarantee funding for Horizon Europe MSCA Postdoctoral Fellowships (EP/Z002206/1), UKRI MRC Research Grant, TFS Research Grants (MR/U506710/1), and the UKRI Future Leaders Fellowship (MR/V023799/1). Zhenxuan Zhang was supported by a CSC Scholarship.

\bibliography{ref}

\end{document}

\maketitle

\section*{A. Case Study: GEMA-Score Evaluation on Structured Radiology Report Entities}
\begin{figure*}[h]
\centerline{\includegraphics[width=\columnwidth]{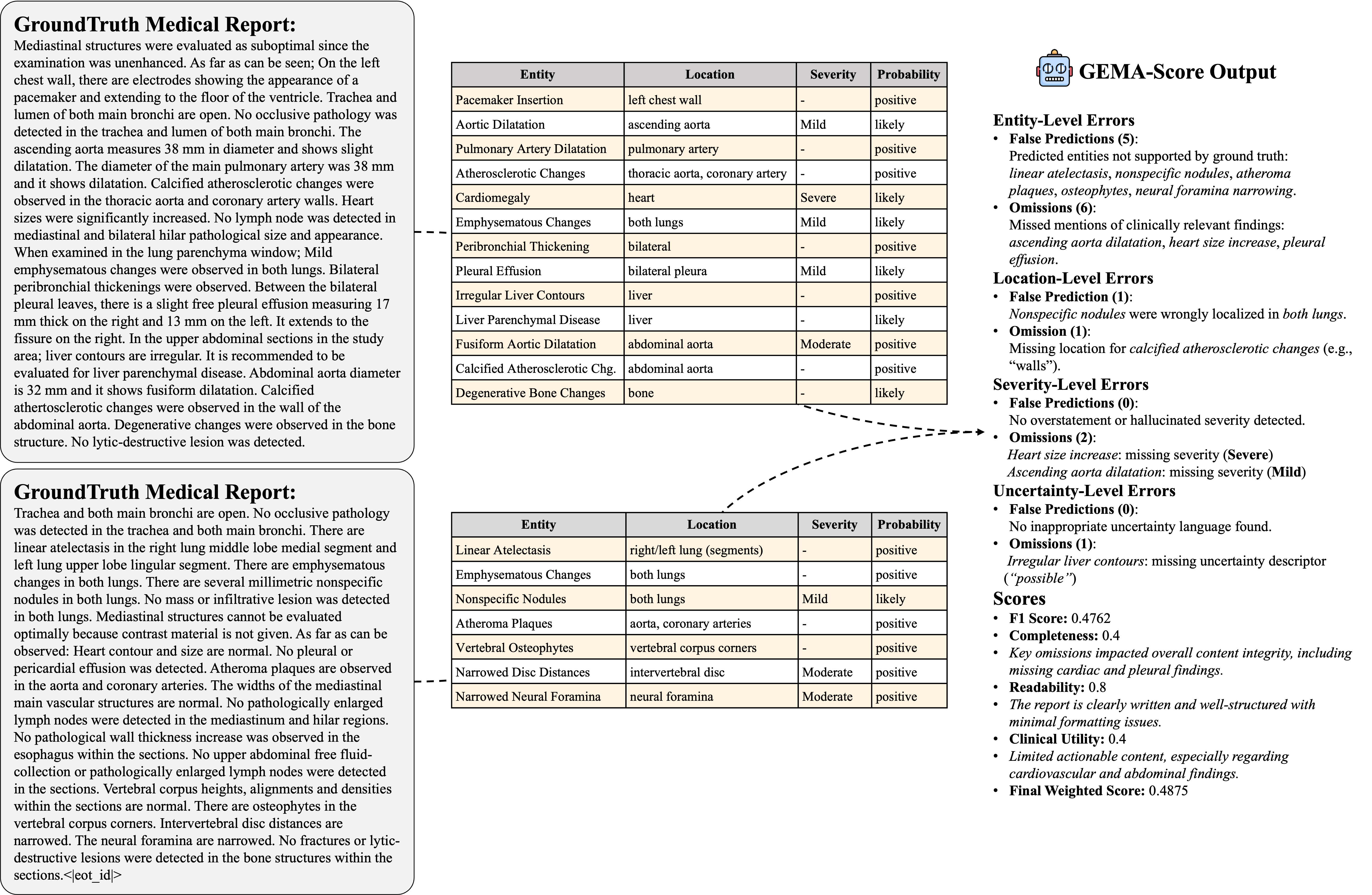}}
\caption{GEMA-Score evaluation on a CT case. The example includes the original report (left), a structured table of extracted entities with their spatial and diagnostic attributes (center), and detailed feedback generated by GEMA-Score (right). The system captures semantic alignment between structured data and complex CT findings.}
\label{fig:sup_1}
\end{figure*}
\begin{figure*}[h]
\centerline{\includegraphics[width=\columnwidth]{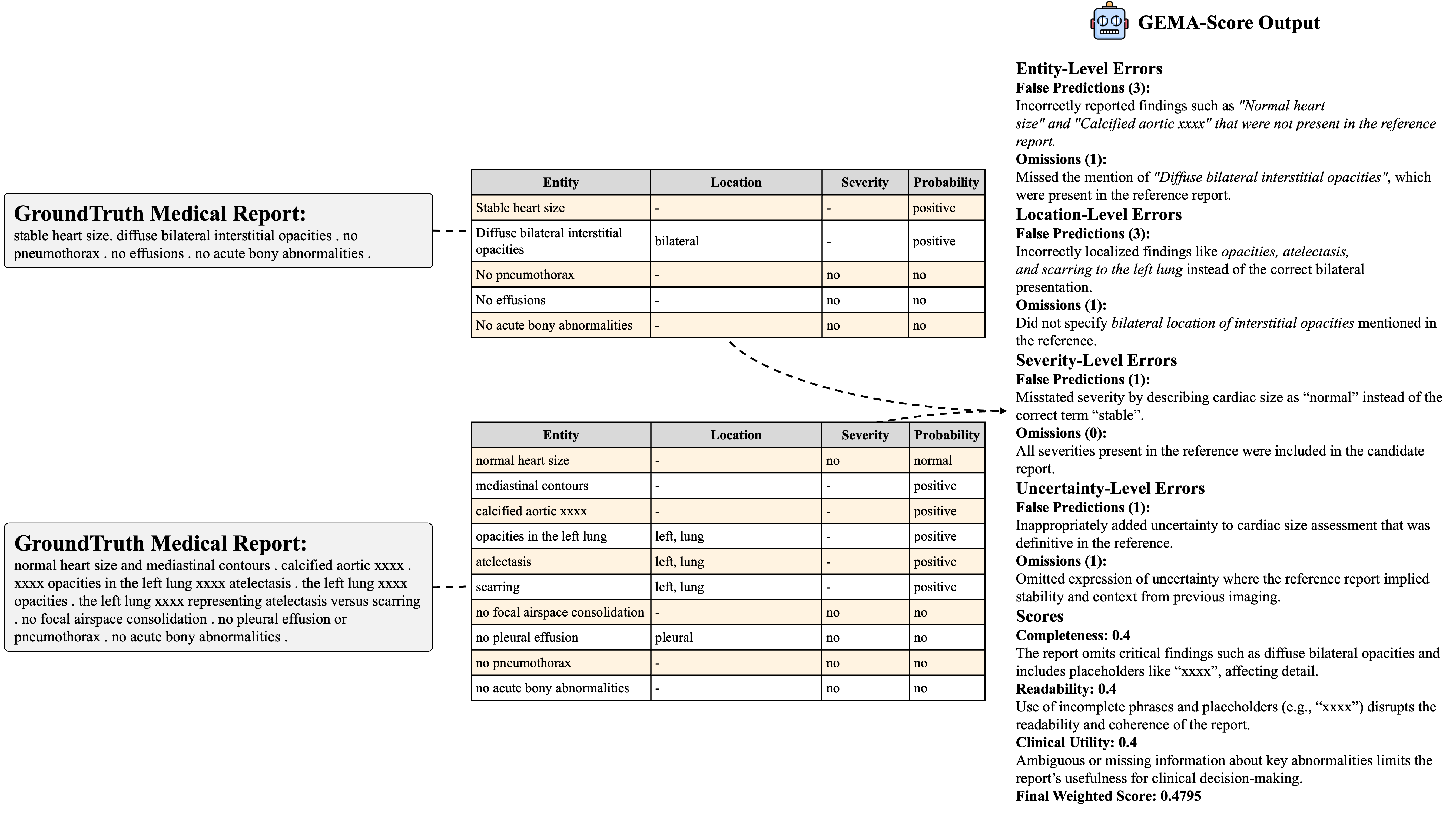}}
\caption{GEMA-Score evaluation on a chest X-ray case. Shown are the radiology report (left), the corresponding entity table (center), and GEMA-Score analysis (right). Compared to CT, X-ray reports often contain less spatial detail; GEMA-Score remains capable of providing fine-grained, explainable assessments for each predicted entity.}
\label{fig:sup_2}
\end{figure*}
To qualitatively assess the effectiveness of GEMA-Score, we present two case studies illustrating its evaluation results. Each case consists of three parts:

\begin{itemize}
  \item \textbf{Left:} the original ground-truth radiology report written by clinicians;
  \item \textbf{Center:} the structured extraction of clinical entities, including their anatomical location, severity (if available), and diagnostic probability;
  \item \textbf{Right:} the GEMA-Score evaluation output for each entity, providing granular, explainable feedback based on semantic alignment between the extracted information and the source report.
\end{itemize}

These examples demonstrate how GEMA-Score effectively captures semantic correctness, uncertainty attribution, and entity-level granularity, offering interpretable assessment of generated or extracted medical content.
\newpage

\section*{B. Distributional Analysis of Structured Evaluation Metrics Across Models}
\begin{figure*}[h!]
\centerline{\includegraphics[width=\columnwidth]{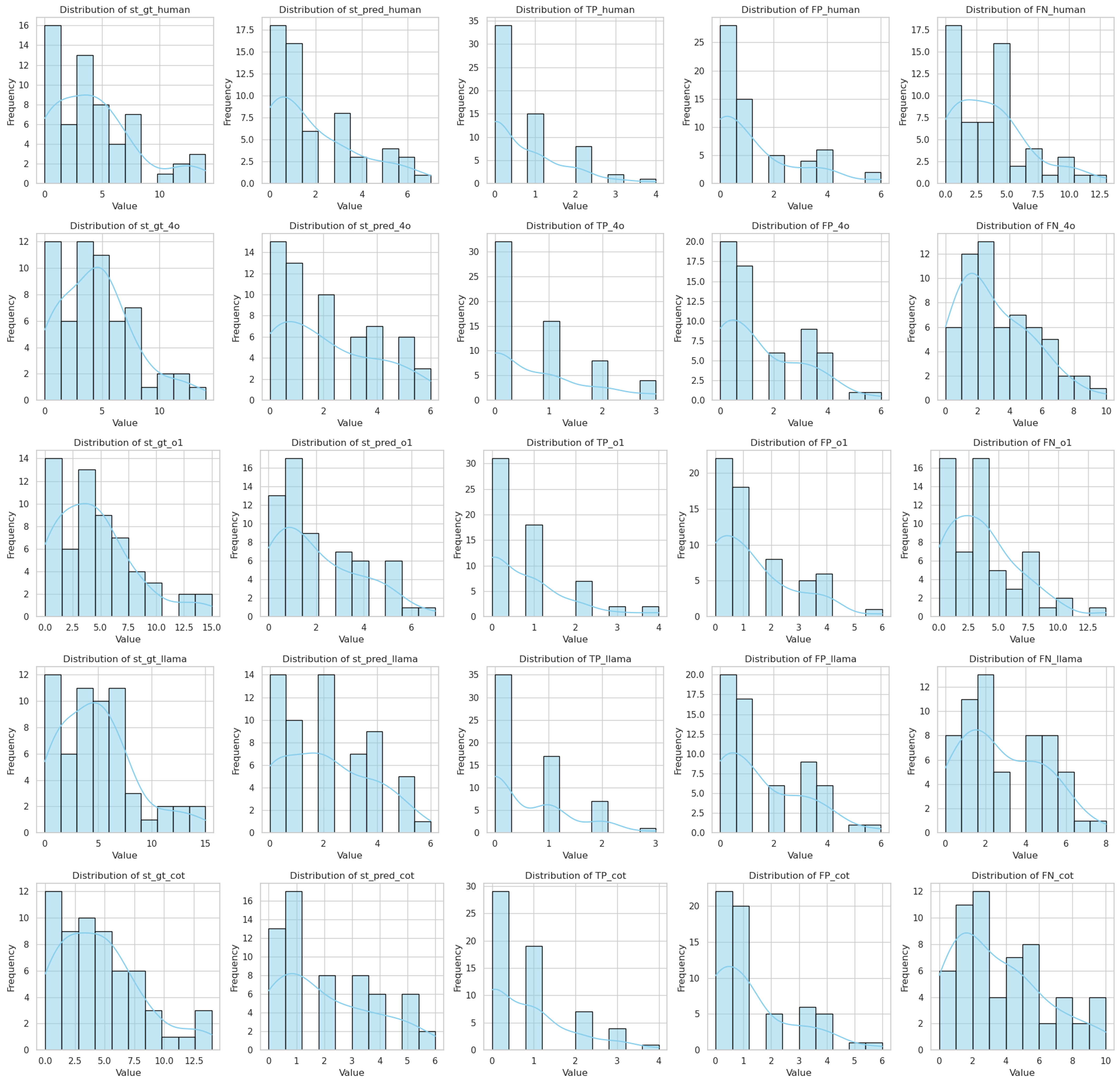}}
\caption{Distributional comparison of evaluation metrics across different sources. Each row corresponds to a source (Human, GPT-4o, o1, LLaMA, CoT), and each column represents one of the five structured metrics: ground-truth count (\texttt{st\_gt}), predicted count (\texttt{st\_pred}), true positives (TP), false positives (FP), and false negatives (FN). Each subplot shows the histogram of sample-wise values with a kernel density estimation curve overlaid. The figure illustrates distinct behavioral patterns across agents in structured entity prediction.}
\label{fig:sup_2}
\end{figure*}

\newpage

\section*{C. Cross-Metric Correlation Analysis Between Human and Models}
To assess the consistency between different models and human annotations on structured evaluation metrics, we compute the Pearson correlation coefficients across five key dimensions: ground-truth entity count (\texttt{st\_gt}), predicted count (\texttt{st\_pred}), true positives (TP), false positives (FP), and false negatives (FN).

Each heatmap illustrates the correlation between the metrics produced by a given model and those from human annotations. High correlation along diagonal elements indicates structural consistency in entity understanding, while off-diagonal values reveal how prediction components (e.g., FP vs. TP) may diverge from human behavior.

This analysis provides insight into which models are more aligned with human-level extraction behavior, and which types of errors (e.g., over-detection or under-detection) they are prone to.
\begin{figure*}[h]
\centerline{\includegraphics[width=\columnwidth]{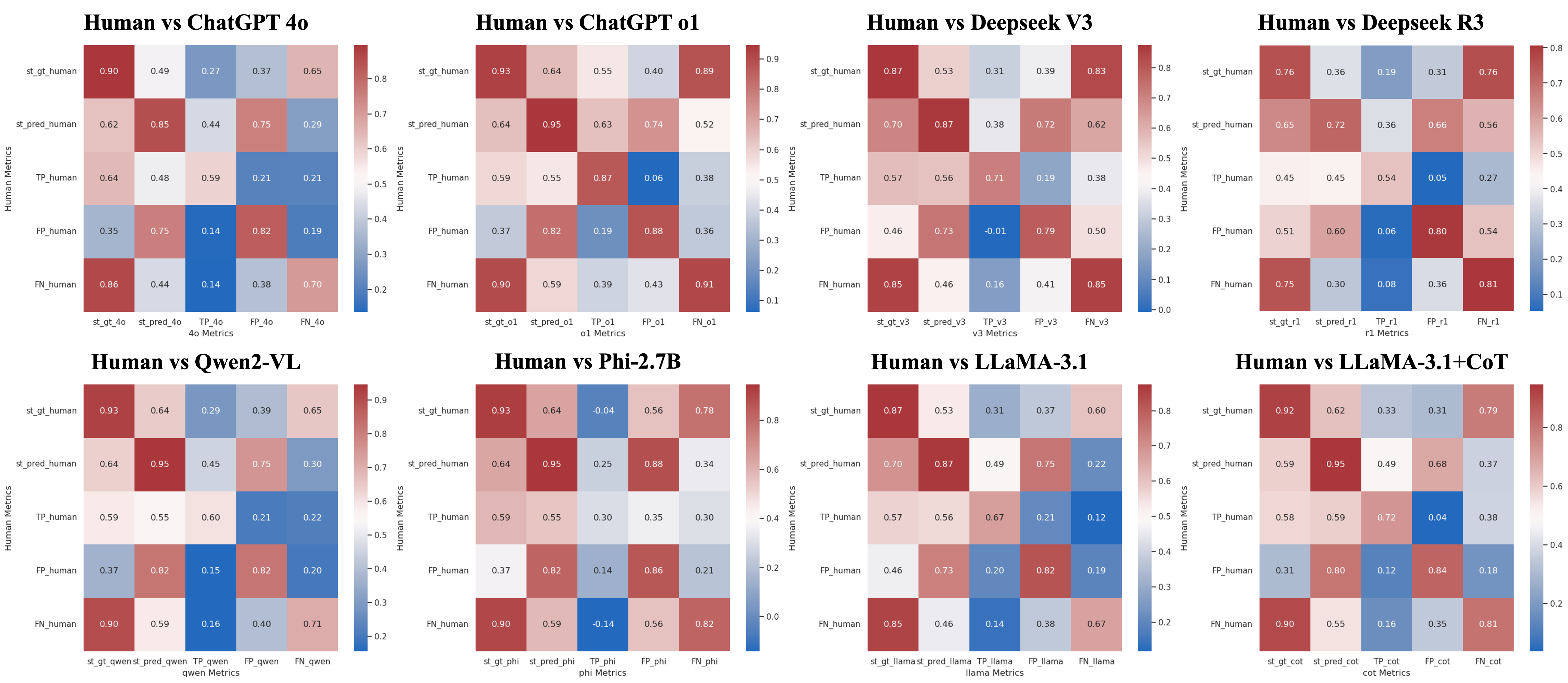}}
\caption{Cross-metric Pearson correlation between human annotations and model outputs across structured entity evaluation metrics. Each heatmap corresponds to one model (GPT-4o, o1, v3, r1, Qwen, Phi, LLaMA, and CoT), comparing five metrics: \texttt{st\_gt}, \texttt{st\_pred}, TP, FP, and FN. Stronger diagonal correlation values indicate higher structural alignment with human judgment, while lower or negative off-diagonal values may reflect systematic prediction biases.}
\label{fig:sup_3}
\end{figure*}
\newpage

\section*{D. Comprehensive Metric Benchmarking on X-ray and CT Reports}

\begin{table*}[h]
\centering
\resizebox{\columnwidth}{!}{
\begin{tabular}{@{}l|l|ccc|ccc|c|c|c}
% \multicolumn{11}{l}{\large{\textbf{(a) Existing Evaluation Score:}}}\\

\hline
\multirow{2}{*}{\textbf{Category}} & \multirow{2}{*}{\textbf{Model}} & \multicolumn{3}{c|}{\textbf{Overlap-Based}} & \multicolumn{3}{c|}{\textbf{BERT-Based}} & \multicolumn{1}{c|}{\textbf{NER-F1}}& \multicolumn{1}{c|}{\textbf{LLM-based}} & \multirow{2}{*}{\textbf{ Gema-Score}} \\ 
\cline{3-10} 
&  & 
\makecell{BLEU-1 } & 
\makecell{ROUGE } & 
\makecell{METEOR } & 
\makecell{BERTScore } & 
\makecell{BLUERT} & 
\makecell{MOVERScore } & 
\makecell{RadGraph-F1} & 
\makecell{Green} & 
 \\ 
\hline
\multirow{6}{*}{\textbf{AMRG Model}} 
& R2GEN       & 0.326 & 0.276 & 0.159 & 0.933 & 0.630 & 0.340 & 0.137 & 0.276 & 0.534 \\
& R2GEN-CMN   & 0.340 & 0.284 & 0.168 & 0.934 & 0.637 & 0.334 & 0.148 & 0.295 & 0.530 \\
& RGRG        & 0.168 & 0.146 & 0.158 & 0.907 & 0.538 & 0.370 & 0.152 & 0.246 & 0.553 \\
& CheXagent   & 0.155 & 0.163 & 0.186 & 0.904 & 0.529 & 0.359 & 0.165 & 0.263 & 0.537 \\
& RadFM       & 0.067 & 0.081 & 0.049 & 0.787 & 0.427 & 0.535 & 0.001 & 0.000 & 0.201 \\
& Mini-GPT    & 0.054 & 0.078 & 0.031 & 0.753 & 0.404 & 0.577 & 0.012 & 0.007 & 0.233 \\
\hline
\multirow{3}{*}{\makecell{\textbf{Open-source}\\\textbf{Model}}} 
& InternVL2             & 0.040 & 0.085 & 0.043 & 0.887 & 0.464 & 0.455 & 0.062 & 0.119 & 0.487 \\ 
& LLAVA                 & 0.070 & 0.084 & 0.043 & 0.784 & 0.426 & 0.534 & 0.016 & 0.024 & 0.280 \\
& HuatuoGPT-o1          & 0.081 & 0.089 & 0.057 & 0.851 & 0.466 & 0.497 & 0.062 & 0.089 & 0.451 \\
\hline
\hline
\end{tabular}}
\caption{Benchmark comparison of multiple models on chest X-ray report generation, using traditional NLG metrics (BLEU, ROUGE, METEOR), semantic similarity metrics (BERTScore, BLUERT, MOVERScore), RadGraph-based entity extraction F1, LLM-based evaluation (Green), and the proposed GEMA Score.}
\end{table*}

% \begin{table*}[ht]
% \centering
% \footnotesize
% \setlength{\tabcolsep}{4pt}
% \resizebox{\columnwidth}{!}{
% \begin{tabular}{l|cccccc|ccc|ccc|ccc}
% \toprule
% & \multicolumn{6}{c|}{\textbf{NLG Metrics}} & \multicolumn{3}{c|}{\textbf{CE Weighted}} & \multicolumn{3}{c|}{\textbf{CE Sample}} & \multicolumn{3}{c}{\textbf{GEMA Score}} \\
% \textbf{Method} & \textbf{BLEU-1} & \textbf{BLEU-2} & \textbf{BLEU-3} & \textbf{BLEU-4} & \textbf{METEOR} & \textbf{ROUGE-L} & 
% \textbf{P} & \textbf{R} & \textbf{F1} & 
% \textbf{P} & \textbf{R} & \textbf{F1} & 
% \textbf{P} & \textbf{R} & \textbf{F1} \\
% \midrule

% M3D             & 44.95 & 34.36 & 27.83 & 22.98 & 40.75 & 39.43 & 47.06 & 28.54 & 33.13 & 35.04 & 23.22 & 25.21 & 32.59 & 19.30 & 24.24 \\
% Reg2RG & 44.89 & 33.43 & 26.42 & 21.08 & 38.25 & 24.41 & 26.19 & 11.06 & 14.51 & 19.81 & 11.50 & 12.19 & 16.36 & 11.64 & 13.60 \\
% CT-CTCHAT-8B & 38.12 & 27.28 & 20.65 & 15.80 & 34.31 & 41.93 & 27.29 & 34.08 & 23.27 & 24.39 & 28.10 & 23.10 & 18.55 & 23.71 & 20.81 \\
% % CT-CTCHAT-70B         & 36.58 & 26.98 & 21.26 & 17.17 & 34.61 & 24.25 & 27.78 & 20.05 & 18.26 & 22.97 & 16.32 & 16.44 & 18.93 & 20.46 & 19.63 \\
% \bottomrule
% \end{tabular}}
% \caption{Performance benchmark of selected models on CT-based report generation using the same suite of metrics as the X-ray benchmark. The results highlight domain-specific differences and further validate the robustness of GEMA Score in assessing structured clinical accuracy.}

% \label{tab:main_metrics}
% \end{table*}
To comprehensively evaluate the performance of existing automatic report generation and grounding models, we benchmark multiple methods on both chest X-ray and CT datasets across a diverse set of evaluation metrics.

For each method, we report scores from traditional natural language generation (NLG) metrics (BLEU, METEOR, ROUGE), semantic similarity metrics (BERTScore, BLUERT, MOVERScore), clinical entity-based metrics (RadGraph-F1), as well as large language model (LLM)-based metrics (Green) and our proposed GEMA Score. These metrics reflect different aspects of clinical report quality, including surface similarity, semantic alignment, and structured entity correctness.

The first table summarizes the performance of representative models on the X-ray domain, while the second table shows results on CT reports, highlighting the distinct challenges and variability across modalities.